\documentclass[10pt, a4paper]{article}
\usepackage{lrec}
\usepackage{multibib}
\newcites{languageresource}{Language Resources}
\usepackage{graphicx}
\usepackage{tabularx}
\usepackage{soul}
\usepackage{CJK}
\usepackage{amsmath}
\usepackage{xstring}
\usepackage{bm}
\usepackage{epstopdf}

\usepackage{hyperref}

\title{A Chinese Dataset with Negative Full Forms for General Abbreviation Prediction}

\name{Yi Zhang, Xu Sun}

\address{MOE Key Laboratory of Computational Linguistics, Peking University\\ School of Electronics Engineering and Computer Science, Peking University\\
{zhangyi16, xusun}@pku.edu.cn
}

\abstract{
	Abbreviation is a common phenomenon across languages, especially in Chinese. In most cases, if an expression can be abbreviated, its
abbreviation is used more often than its fully expanded forms, since people tend to convey information in a most concise way. For various
language processing tasks, abbreviation is an obstacle to improving the performance, as the textual form of an abbreviation does not
express useful information, unless it's expanded to the full form. Abbreviation prediction means associating the fully expanded forms with
their abbreviations. However, due to the deficiency in the abbreviation corpora, such a task is limited in current studies, especially
considering general abbreviation prediction should also include those full form expressions that do not have valid abbreviations, namely
the negative full forms (NFFs). Corpora incorporating negative full forms for general abbreviation prediction are few in number. In order to
promote the research in this area, we build a dataset for general  Chinese abbreviation prediction, which needs a few preprocessing steps, and
evaluate several different models on the built dataset. The dataset is available at \url{https://github.com/lancopku/Chinese-abbreviation-dataset}.
\newline  }

\begin{document}
	
	\maketitleabstract
	
\section{Introduction}
Abbreviation processing mainly consists of three tasks, that is, abbreviation expansion, abbreviation recognition, and abbreviation prediction. Expanding the short form of an expression to its full form is called abbreviation expansion. Extracting the short form and full form pairs from the context is called abbreviation recognition. Abbreviation prediction refers to predicting the short form of an expression according to its full form. In this paper, we focus on the last task, i.e., abbreviation prediction. Abbreviation prediction plays an important role in various language processing tasks, because accurate abbreviation prediction will help improve performance. \newcite{sun2009robust} shows that better abbreviation prediction will help improve the performance of abbreviation recognition. Abbreviation prediction also benefits other tasks. For example, in an information retrieval (IR) system, a large number of the web pages contain only abbreviations. It will be helpful if we can estimate abbreviations of a query,  because successful abbreviation prediction may improve the recall of IR systems as \newcite{sun2013generalized} showed. In addition, \newcite{yang2012vocabulary} showed that Chinese abbreviation prediction can improve voice-based search quality. 

\begin{figure}[ht]
	\begin{center}
		\includegraphics[width = 7cm]{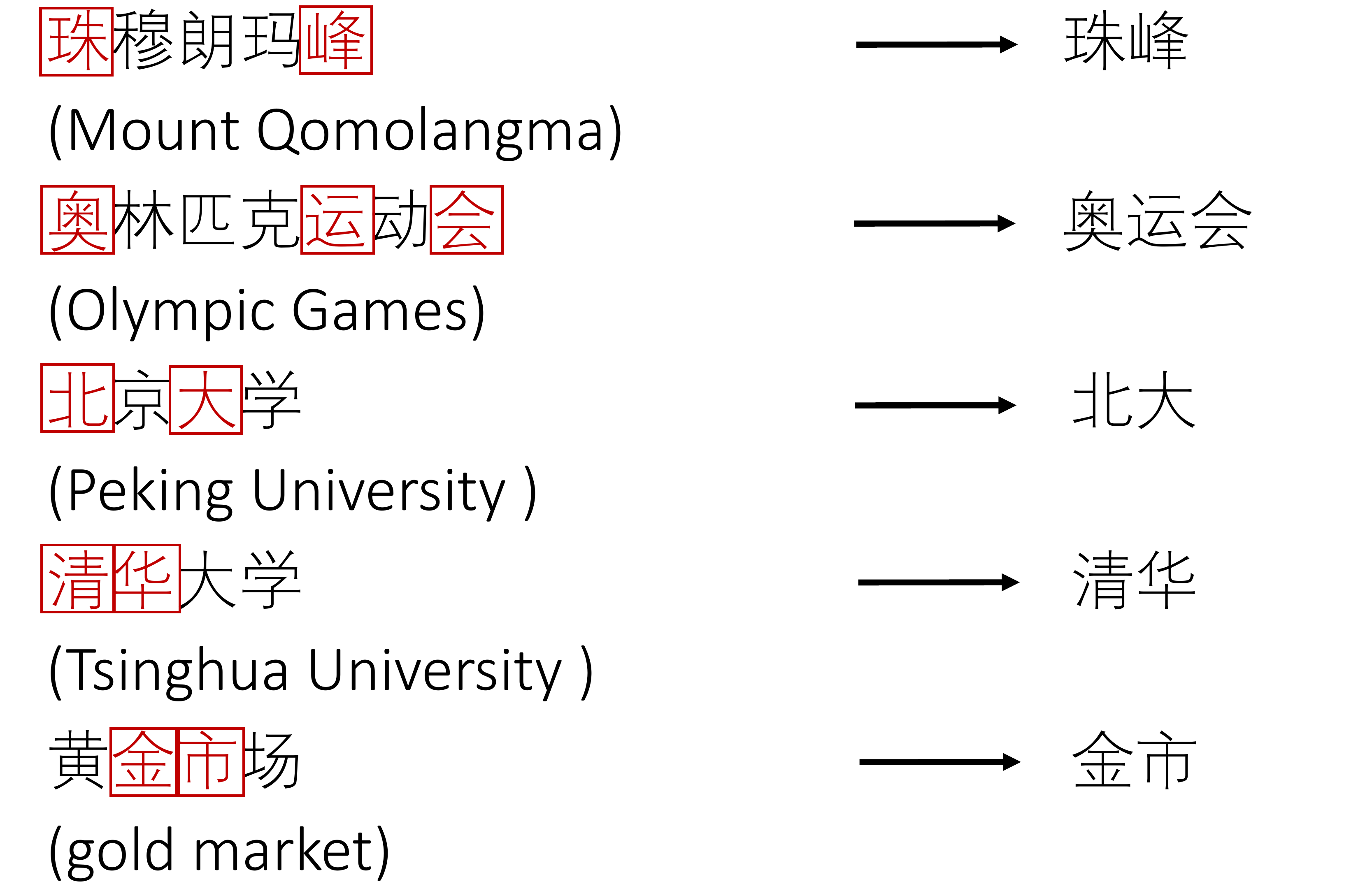}
	\end{center}
	\caption{Different cases of generating abbreviations}\label{fig1}
\end{figure}

English abbreviations are usually formed as acronyms. Studies for English abbreviation proposed various heuristics for abbreviation prediction. For example, use of initials, capital letters, syllable boundaries, stop words, etc. These studies performed well for English abbreviations. While Chinese abbreviations are quite different from English ones. \newcite{yang2012vocabulary} showed that Chinese abbreviations are usually generated by three methods, reduction, elimination, and generalization. Characters are selected from the expanded full name to form the abbreviation. However, there are no general rules to convert a complete term into an abbreviation.  As shown in Figure~\ref{fig1}, an abbreviation may be generated using the first character and the last character. Sometimes, characters in the middle can be included while the last abbreviation takes the first two characters of the words. However, it is not necessary for Chinese abbreviations to take the first characters of words. They frequently take non-initial characters, like the last example in Figure~\ref{fig1}. Chinese abbreviations are derived via a customary lexical process. Native speakers may associate a fully expanded term with its abbreviation by some intuition. But the process can not be adequately explained by any linguistic theory: \newcite{chang2004preliminary} and \newcite{chang2006mining}.

Besides the irregularity of abbreviating phrases and terms,  another main problem is caused by negative full forms. A word annotated with a negative full form means the word has no abbreviation at all. We usually recognize abbreviations or make abbreviation predictions in text.  Unfortunately, NFFs take up a large portion of Chinese words or phrases in the real world. With the strong noise, distinguishing the full forms with valid abbreviations is more difficult. This undoubtedly increases the difficulty of abbreviation prediction.  

Many approaches have been proposed in the post studies. \newcite{sun2008predicting} employed Support Vector Regression (SVR) for scoring abbreviation candidates. This method outperforms the hidden Markov model (HMM) in abbreviation prediction. \newcite{yang2009automatic} proposed to formulate abbreviation generation as a character tagging problem and conditional random field (CRF) then can be used as the tagging model. \newcite{sun2009robust} combined latent variable model and global information to predict abbreviations. \newcite{zhang2016generating} used a recurrent neural networks to predict abbreviations for Chinese named entities.

However, most studies of abbreviation predictions focus on positive full form, which means a word has a valid abbreviation.  Apparently,  this implicit lab assumption is not practical. Nonetheless, we barely see studies that consider NFFs. One of the main reasons is the shortage of abbreviation prediction data with NFFs, which is one of the main issues this work tries to solve.

Apart from the annotation of a dataset with NFFs, we also conduct a few preprocessing steps to facilitate the usage of the dataset. Chinese does not insert spaces between words or word forms that undergo morphological alternations. Hence, most of the Chinese natural language processing methods assume a Chinese word segmenter is used in a preprocessing step to produce word-segmented Chinese sentences as inputs. There is no exception for abbreviation prediction. Given original texts, we should first recognize the boundaries of words.   After segmentation,  we annotate the part-of-speech information of phrases and terms. Because the part-of-speech information can serve as features to help make abbreviation prediction.

This paper details how the dataset is created and evaluates some frequently used models on the abbreviation prediction task.

\section{Dataset}

\subsection{Considerations}

\textbf{Commonness}
Our intention is to build a dataset with NFFs, so it can be widely used for general abbreviation prediction. This requires that dataset contains most frequently-used full forms regardless of whether or not the form has valid abbreviations. The data sources should be reliable and accredited. Thus, we extract long phrases and terms in popular Chinese natural language processing corpora, which includes People's Daily corpora and SIGHAN word segmentation corpora.

\textbf{Usability}
We also provide assisting information that is helpful for the abbreviation task in our dataset. Most existing methods treat abbreviation prediction as a sequence labeling problem. To make better tag predictions of characters, we usually need to extract some features.The word segmentation information and part-of-speech information are most commonly used features. Unlike English, the smallest Chinese unit is a character rather than a word. There are no explicit boundaries between Chinese words. Since a full form usually can be segmented into several words and abbreviations often take characters from these words, segmentation information is most useful for abbreviation prediction. Another annotation is part-of-speech information. Many language processing tasks take part-of-speech information as features, including abbreviation prediction. The choice of characters which are used to form abbreviations may be related to their part-of-speech information. 

\textbf{Representativeness}

\begin{figure}[ht]
	\begin{center}
		\includegraphics[width = 7cm]{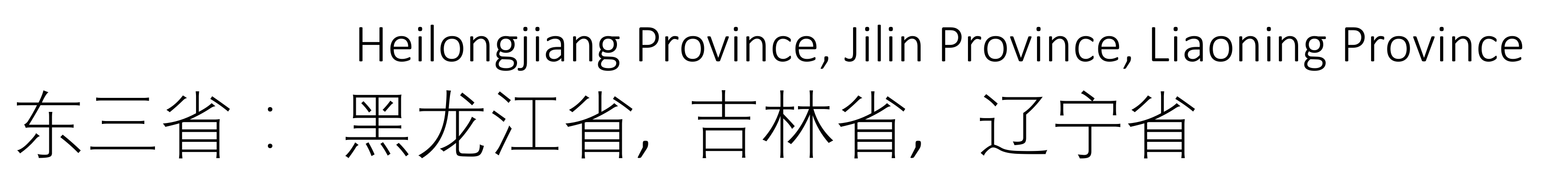}
	\end{center}
	\caption{A special case for abbreviation.}\label{fig4}
\end{figure}

As the dataset should be representative of the common construction of abbreviations, we do not include special and irregular abbreviations, which include words outside the full form. As shown in Figure~\ref{fig4}, ``\begin{CJK*}{UTF8}{gbsn}
东三省
\end{CJK*} 
''represents three provinces of China. It is a special type of abbreviations, since an abbreviation could represent several different terms. Without some background knowledge, what the abbreviation stands for can not be understood. Sometimes the characters of the abbreviation are not taken from original characters of the full form and the sequence labeling method is no longer applicable for this case. This kind of ``abbreviation'' is more like a general name for some terms. We do not include these special abbreviations in our dataset.

\subsection{Data Source}

\begin{figure}[ht]
	\begin{center}
		\includegraphics[width = 8cm]{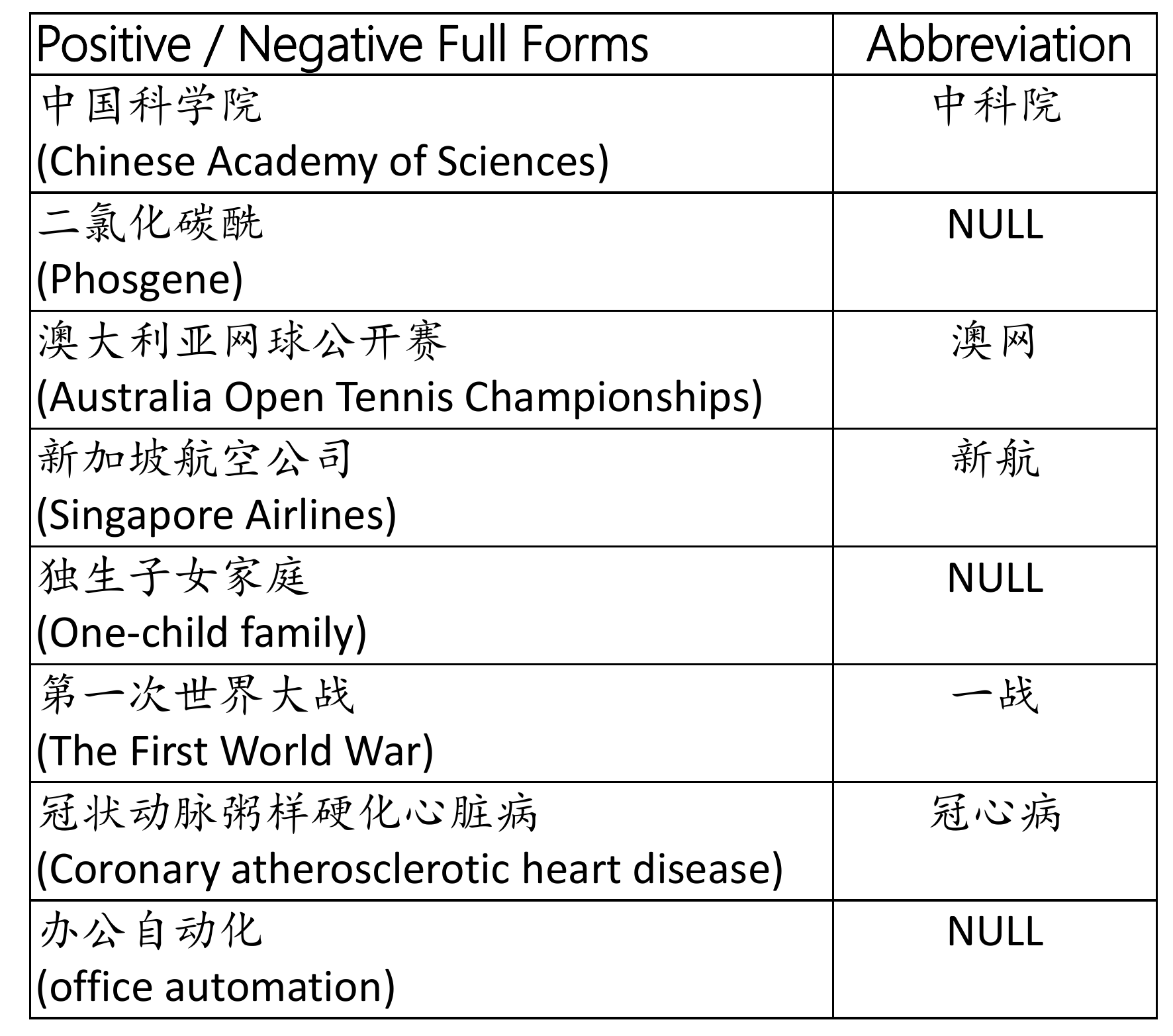}
	\end{center}
	\caption{Samples of the collected data with NFFs.
		The ``NULL'' means no valid abbreviation.}\label{fig2}
\end{figure}

Our text is from People's Daily corpora and SIGHAN word segmentation corpora. We extract the long phrases and terms in the text. Then we classify the collected phrases and terms into two forms. One is the positive full form, which means the phrase or term has a valid abbreviation. Then its abbreviation is annotated.  The other is the negative full form, which means the phrase or term can not be shortened. Their abbreviations are NULL. Samples of the data are shown in Figure~\ref{fig2}.

As mentioned before, we annotate word segmentation information and part-of-speech information for every phrase or term. Word segmentation is a fundamental task in Chinese processing. Many practical Chinese processing applications rely on Chinese word segmentation. Part-of-speech information is often used as features for further prediction. In the general abbreviation prediction task, many full forms that can be shortened are labeled with noun tags. Most methods formulate these tasks as a sequence labeling problem. Various models achieved good performance on these tasks and some open source tools have been published for use. We used ICTCLAS, one of the best Chinese Lexical analyzers, to label the segmentation and part-of-speech information.

\subsection{Statistics}
We build a dataset that is made up of phrases and terms. There are 10,786 full forms in this dataset, including 8,015 positive full forms and 2,661 negative full forms. The phrases contain noun phrases, verb phrases, organization names, location names, and so on. The distribution is shown in Table~\ref{tab2}. For experiments, we randomly sampled 7,551 samples as the training set, 1078 samples as the development set and 2,157 samples as the testing set. We calculate the numbers of the words and characters (including duplicates) in the data. We also count the numbers of distinct words and distinct characters. Then total characters of full forms divided by total entries is the average full form length. The average abbreviation length can be calculated in a similar way.

\begin{table}
	\begin{center}
		\begin{tabular}{l|l}
			\hline
			Full Forms \\
			\hline
			total entries &10,786 \\
			NFFs &2,661        \\
			total words &30,100      \\
			distinct words &8,293      \\
			total characters &60,877      \\
			distinct characters &2,557      \\
			average word length &5.644     \\
			
			\hline
			Abbreviations \\
			\hline
			total characters &23,077      \\
			distinct characters &1,687      \\
			average abbreviation length &2.140     \\
			\hline
		\end{tabular}
	\end{center}
    \caption{The statistics of the data. The results are count for full forms and abbreviations separately.}
			
\end{table}\label{tab1}

\begin{table}[!ht] % table 2
	\begin{center}
		\begin{tabular}{l|l}
			\hline
			Category &Portion(\%)     \\
			\hline
			Noun Phrase &52.01\%  \\
			Organization Name &26.84\% \\
	     	Verb Phrase &13.72\% \\
	     	Location Name &5.28\% \\
	     	Person Name &0.32\% \\
		    Others  &1.80\% \\
		
			\hline
		\end{tabular}
	\end{center}
	\caption{Distribution of the full forms in the data.}\label{tab2}

\end{table}

\begin{table*}[ht] 
	\begin{center}
		\begin{tabular}{|l|l|l|l|}
			\hline
			Method   &Discriminate Acc(\%) &Overall All-Acc(\%)  &Overall Char-Acc(\%) \\
			\hline
			Heuristic System  &73.20  &25.77 &65.79  \\
			Perc              &87.48  &54.89 &87.02  \\
			MEMM              &86.97  &50.16 &85.92  \\
			CRF-ADF           &87.80  &56.69 &87.20  \\
			BLSTM             &91.38  &57.30 &82.01  \\	
			\hline
		\end{tabular}
	\end{center}
	\caption{Results on comparing different methods on generalized abbreviations.}\label{tab3}
	%\vspace{-0.1in}
\end{table*}

\begin{figure}[ht]
	\begin{center}
		\includegraphics[width = 5.5cm]{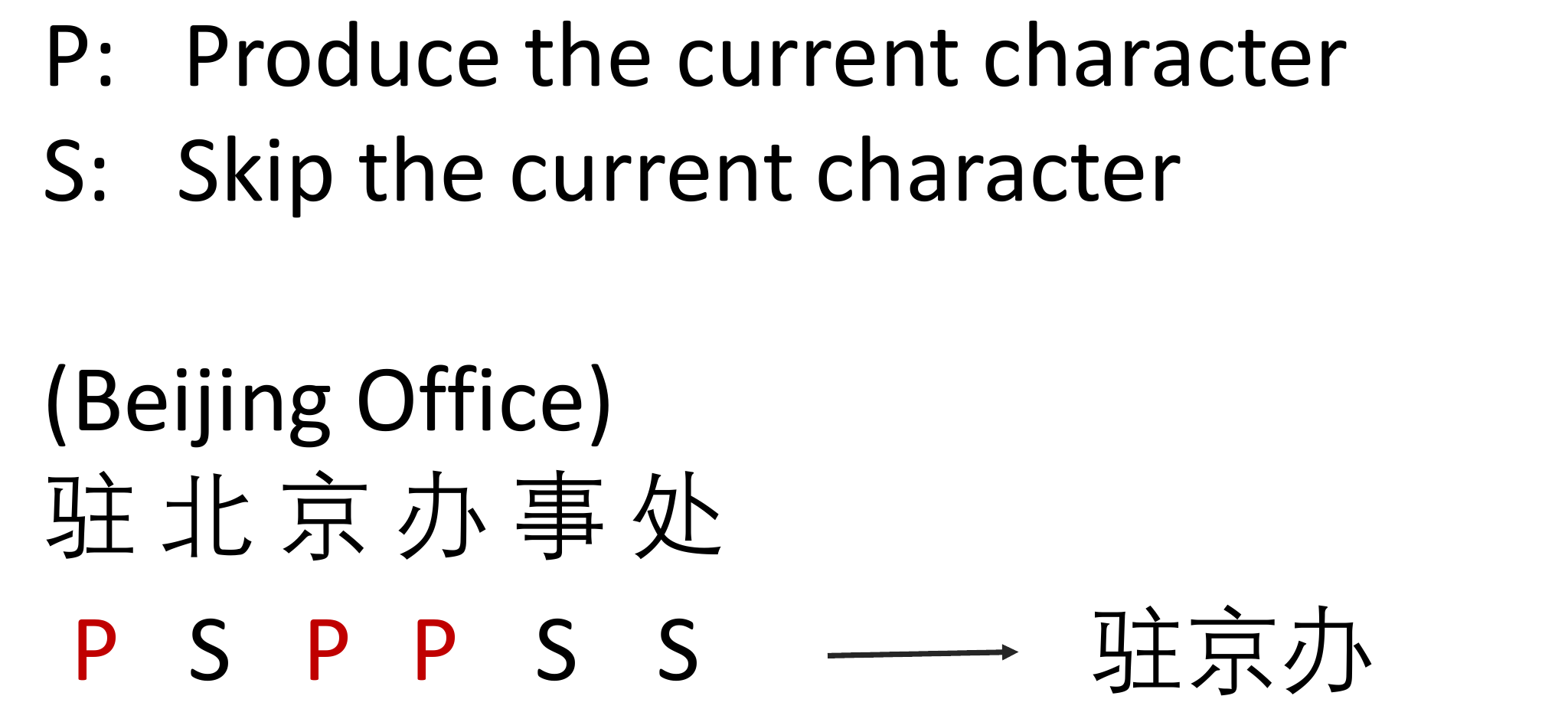}
	\end{center}
	\caption{Chinese abbreviation generation as a sequential labeling problem.}\label{fig3}
\end{figure}

\section{Models}

\subsection{CRF}
\newcite{tsuruoka2005machine} formalized the process of abbreviation prediction as a sequence labeling problem.  Each character in the expanded form is tagged with a label, y $\in$ $\left\{P, S\right\}$, where the label $P$ produces the current character and the label $S$ skips the current character.  In Figure~\ref{fig3}, the abbreviation is generated using the first character, skipping the flowing character and then using the subsequent two characters. Because our task is general abbreviation prediction, we add another label ``$N$'' to the tag set to label the characters in negative full forms.  A number of recent studies have investigated the use of machine learning techniques. Traditional models like MEMM, peceptron and conditional random fields perform well in such sequence labeling tasks. We use the well-known conditional random fields (CRFs) proposed by \newcite{lafferty2001conditional} for sequential labeling. 

We use features as follows:
\begin{itemize}
	\item character feature : Input characters $x_{i-1}$, $x_{i}$ and $x_{i+1}$ 
	\item character bi-gram : The character bigrams starting at $(i-2) \cdots i$.
	\item Numeral: Whether or not the $x_i$ is a numeral.
	\item Organization name suffix: Whether or not the $x_i$ is a suffix of traditional Chinese organization names.
	\item Location name suffix: Whether or not the $x_i$ is a suffix of traditional Chinese location names.
	\item Word segmentation information: After the word segmentation step, whether
	or not the $x_i$ is the beginning character of a word.
	\item Part-of-speech information: The part-of-speech tag information of $x_i$ .
	
\end{itemize}

In our abbreviation prediction task, the input sequence $\bm{x}$ represents characters of a full form and output sequence $\bm{y}$ represents symbolic labels based on abbreviations. The probability is defined as follows:
\begin{equation}
P(\bm{y}|\bm{x},\bm{w}) = \frac{exp[\bm{w}^{T}\bm{f}(\bm{y},\bm{x})]}{\sum_{\forall \bm{y}^{'}}exp[\bm{w}^{T}\bm{f}(\bm{y}^{'},\bm{x})]}
\end{equation}
where $\bm{w}$ is the weight vector and $\bm{f}$ is the mapping function.

Given a training set consists of $n$ labeled sequences ($\bm{x}^{i}$, $\bm{y}^{i}$) for $i=1 \cdots n$, the objective function is:
\begin{equation}
L(w) = \sum_{i=1}^{n}logP(\bm{y}^{i}|\bm{x}^{i},\bm{w})-R(\bm{w})
\end{equation}
where the second term is the $L_2$ regularizer.

\subsection{BLSTM}
As mentioned above, traditional methods depend heavily on features which need to be designed elaborately. Nowadays, more and more researches focus on neural networks, such as recurrent neural networks (RNN), convolutional neural network (CNN) and some variants of RNN. These neural network models can extract features automatically. In natural language processing, traditional RNNs usually take the previous state $h_{t-1}$ and the embedding $x_t$ as the $t$-th input to calculate current state $h_t$. Formally, we have
\begin{equation}
h_t = f(W \cdot x_{t} + V \cdot h_{t-1} + b_h)
\end{equation}
where $W$ and $V$ are weight matrices, respectively. $b_h$ is a bias term and $f$ is a non-linear activation function.

In theory, RNN can keep a memory of previous information. However, it was difficult to train RNNs to capture longterm dependencies because the gradients tend to either vanish or explode. Therefore, some sophisticated variants of RNN were proposed. Long-short term memory units are proposed in \newcite{hochreiter1997long}. This model introduces a gating mechanism, which controls the proportions of information to forget and to pass on to the next time step. Concretely, the LSTM-based recurrent neural network comprises four components: an input gate $i_t$, a forget gate $f_t$, an output gate $o_t$, and a memory cell $c_t$. LSTM memory cell is implemented as following:
\begin{equation}
\begin{split}
&f_t=\sigma(W_{f} \cdot x_t + U_f \cdot h_{t-1} + b_f)\\
&i_t=\sigma(W_i \cdot x_t + U_i \cdot h_{t-1} + bi)\\
&\tilde{C_t}=tanh(W_C \cdot x_t + U_C \cdot h_{t-1} + b_C) \\
&C_t=f_t \otimes C_{t-1} + i_t \otimes \tilde{C_t}\\
&o_t=\sigma(W_o \cdot x_t + U_o \cdot h_{t-1} + b_o)\\
&h_t=o_t\otimes tanh(C_t)
\end{split}
\end{equation}
LSTM can solve the long-distance dependencies problem to some extent. However, the LSTM's hidden
state $h_t$ takes information only from the past, knowing nothing about the future. An elegant solution
whose effectiveness has been proven by previous work \cite{dyer2015transition} is bi-directional LSTM(BLSTM). The basic idea is to present each sequence forwards and backwards to two separate
hidden states to capture past and future information, respectively. Then the two hidden states are concatenated to form the final output. In this paper, we employ a bi-directional LSTM, which could capture the contextual information of the current input, to predict the abbreviations of full terms.  Since we give a specific segmentation tag and a pos tag for every character,  each segmentation tag and pos tag can be mapped to a real-valued vector by looking up their own embedding tables.  These embeddings and character embeddings are all initialized randomly. At current time-step t, the character embedding, segmentation tag embedding and pos tag embedding are concatenated as the input $x_t$. Embeddings of segmentation tag and pos tag are both 20-dimensional. Character embedding is 50-dimensional.  The hidden layer size of BLSTM is 200, 100 for forward LSTM and 100 for backward LSTM. 

\section{Evaluation}
\subsection{Evaluation Metrics}
For evaluating abbreviation prediction quality, the
systems are evaluated using the following two
metrics:

\textbf{All-match accuracy (All-Acc)}: The number of correct outputs (i.e., label strings) generated by the system divided by the total number of full forms in the test set.

\textbf{Character accuracy (Char-Acc)}: The number of correct labels (i.e., a classification on a character) generated by the system divided by the total number of characters in the test set.

\subsection{Simple Heuristic Baseline System}
The simple heuristic system means always choosing initial characters of words in the segmented full form. This is because the most natural abbreviating heuristic is to produce the first character of each word in the original full form. This is just the simplest baseline.
\subsection{Evaluation}
To study the performance of other machine learning models, we also implement other well known
sequential labeling models, including maximum entropy Markov models (MEMMs) \cite{mccallum2000maximum} and averaged perceptrons (Perc) \cite{collins2002discriminative}. Besides these traditional models, we also implement a bidirectional LSTM(BLSTM) to evaluate the performance of neural networks on this task.

The experimental results are shown in Table~\ref{tab3}. In the table, the overall accuracy is most important
and it means the final accuracy achieved by the systems in generalized abbreviation prediction with NFFs. For the completeness of experimental information, we also show the discriminate accuracy. The discriminate accuracy checks the accuracy of discriminating positive and negative full forms, without comparing the generated abbreviations with the gold-standard abbreviations.
The CRF model outperforms the MEMM and averaged perceptron models. The CRF model achieves best overall 
character accuracy. BLSTM outperforms other models in both discriminate accuracy and all-match accuracy. 
However, training a neural network always needs a large amount of data. With a dataset that is not so large,
the ability of a neural network may be limited.

\section{Conlusions and Future Work}
This paper proposes a novel abbreviation prediction dataset with NFFs. Different machine learning methods are evaluated on this general abbreviation task. LSTM shows competitive performance in this task. However, neural networks usually need large data for training. The related corpora are not sufficient and researches for general abbreviation prediction using neural networks are encouraged.

\section{Acknowledgements}
This work was supported in part by National Natural Science Foundation of China (No. 61673028), National High
Technology Research and Development Program of China
(863 Program, No. 2015AA015404), and an Okawa Research
Grant (2016). Email correspondence to Xu Sun.

\section{Bibliographical References}

\nocite{DBLP:journals/talip/SunOTW13,DBLP:conf/emnlp/ZhangWS14, DBLP:conf/emnlp/ZhanglWS14, DBLP:conf/iccpol/SunW06}
\bibliography{bib}
\bibliographystyle{lrec}

\end{document}